\documentclass[runningheads]{llncs}

\usepackage{eccv}

\usepackage{eccvabbrv}

\usepackage{graphicx}
\usepackage{booktabs}

\usepackage[accsupp]{axessibility}  % Improves PDF readability for those with disabilities.

\usepackage{hyperref}

\usepackage{orcidlink}

\usepackage{epsfig}

\usepackage{amsmath}
\usepackage{amssymb}
\usepackage{booktabs}
\usepackage{caption}
\usepackage{subcaption}
\usepackage{graphbox} %loads graphicx package
\usepackage{color, colortbl}

\usepackage{footnote}
\usepackage{threeparttable}

\usepackage{enumitem}
\DeclareMathOperator*{\argmax}{arg\,max}

\usepackage{booktabs}
\usepackage{caption}
\usepackage{subcaption}

\usepackage{algorithm}
\usepackage{amsmath}

\DeclareMathOperator{\logdet}{\log\det}

\usepackage{algpseudocode}

\usepackage{pifont}% http://ctan.org/pkg/pifont

\newcommand{\cmark}{\ding{51}}%
\newcommand{\xmark}{\ding{55}}%

\makeatletter
\newcommand*{\rom}[1]{\expandafter\@slowromancap\romannumeral #1@}
\makeatother

\begin{document}

\title{GradAlign for Training-free Model Performance Inference} 

\titlerunning{GradAlign}

\author{Yuxuan Li\inst{1}  \and
Yunhui Guo\inst{2}}

\authorrunning{Y.~Li et al.}

\institute{Harbin Institute of Technology, China  \and
The University of Texas at Dallas, Richardson, USA\\
\email{lyxzcx@outlook.com, yunhui.guo@utdallas.edu}
}

\maketitle

\begin{abstract}
  Architecture plays an important role in deciding the performance of deep neural networks. However, the search for the optimal architecture is often hindered by the vast search space, making it a time-intensive process. Recently, a novel approach known as training-free neural architecture search (NAS) has emerged, aiming to discover the ideal architecture without necessitating extensive training. Training-free NAS leverages various indicators for architecture selection, including metrics such as the count of linear regions, the density of per-sample losses, and the stability of the finite-width Neural Tangent Kernel (NTK) matrix. Despite the competitive empirical performance of current training-free NAS techniques, they suffer from certain limitations, including inconsistent performance and a lack of deep understanding. In this paper, we introduce GradAlign, a simple yet effective method designed for inferring model performance without the need for training. At its core, GradAlign quantifies the extent of conflicts within per-sample gradients during initialization, as substantial conflicts hinder model convergence and ultimately result in worse performance. We evaluate GradAlign against established training-free NAS methods using standard NAS benchmarks, showing a better overall performance. Moreover, we show that the widely adopted metric of linear region count may not suffice as a dependable criterion for selecting network architectures during at initialization.
  \keywords{Neural Architecture Search \and Deep Learning Theory, Neural Architectures}
\end{abstract}

\vspace{-0.3cm}
\section{Introduction}
\vspace{-0.2cm}
\label{sec:intro}

One of the crucial elements that contribute to the effectiveness of deep neural networks (DNNs) is the careful design of their architectures. For instance, the ResNet architecture, as proposed in \cite{he2016deep}, utilizes skip-connections to train extremely deep neural networks. On the other hand, the U-Net architecture, as proposed in \cite{ronneberger2015u}, features a unique design that incorporates both a contracting path and a symmetrical expanding path for biomedical image segmentation. However, the cost of designing such architectures is significant, as it typically necessitates hundreds or even thousands of GPU hours to create a high-performing network architecture. 

Recently, Neural Architecture Search (NAS) \cite{zoph2016neural,liu2018darts,tan2019mnasnet,radosavovic2020designing,xu2019pc,li2020random,guo2020single,xie2019exploring} has emerged as an effective solution for reducing human effort in the design of network architectures. The primary objective of NAS is to efficiently identify optimal network structures by assessing the performance of various models using a validation set. Various NAS methods have been proposed to improve the target architecture performance and reduce search time \cite{li2020random,guo2020single,xie2019exploring}. Yet, it is still a time-costly process to find the best architecture given the large search space \cite{zoph2016neural,radosavovic2020designing}.

Rather than relying on a trial-and-error process to validate the performance of an architecture, training-free NAS \cite{mellor2021neural} has recently been proposed as an effective means of identifying the optimal architecture at the initialization stage of the network. The key insight behind this approach is that a random network at initialization can already provide indications of its potential performance on the target task. There are several ways to score networks at initialization. For instance, NASWOT \cite{mellor2021neural} leverages the diversity of activation patterns to rank different network architectures based on their ability to produce a higher number of linear regions, while TE-NAS \cite{chen2021neural} evaluates model performance by assessing the network's expressivity (based on the number of activation patterns for a given dataset) and trainability (based on the condition number of the finite-width NTK matrix \cite{jacot2018neural}). More recently, GradSign \cite{zhang2021gradsign} was introduced as a method to select the optimal architecture by leveraging the density of sample-wise loss minima. This method assumes that the denser the sample-wise loss minima, the lower the global minimum for all samples. Despite the remarkable success of training-free NAS and its ability to achieve comparable results with training-based NAS methods, these approaches still encounter challenges such as inconsistent performance improvements and a lack of deeper understanding \cite{colin2022adeeperlook}.

In this paper, we propose GradAlign, a simple yet effective approach for inferring model performance at initialization. A useful insight in existing deep learning literature shows that it is difficult to achieve competitive performance with a network that is harder to optimize \cite{yang2017mean}. In TE-NAS \cite{chen2021neural}, the NTK matrix is leveraged to quantify the optimization complexity of neural networks. Nonetheless, it is important to note that the applicability of NTK theory is only valid for infinitely wide neural networks, indicating that its utility in characterizing the behavior of finite neural architectures may not be reliable \cite{lee2019wide,seleznova2022analyzing}. Motivated by the common practice of training neural networks through gradient-based optimization techniques, we aim to understand conditions under which a network becomes more amenable to optimization, thereby facilitating faster convergence. In particular, we decompose the loss computed by the network into per-example basis and evaluate the gradient for each example separately. We theoretically show that if there is a high interference between per-sample gradients, then the network can converge slower. GradAlign employs two strategies to quantify such interference: one involves aligning per-sample gradients with the average gradients, while the other entails computing the determinant of the gram matrix derived from the per-sample gradients. We conduct a comparative analysis between GradAlign and established training-free NAS techniques using widely recognized NAS benchmarks, namely NAS201 \cite{dong2020bench}, NAS101 \cite{ying2019bench}, and NDS \cite{2019On}. The results show that GradAlign generally outperforms existing training-free NAS methods.
\\
\vspace{-0.3cm}
\noindent \textbf{Contributions.}
\begin{itemize}
    
    \item We theoretically show that conflicting per-sample gradients can be lead to slower model convergence. Leveraging this insight, we introduce a novel approach named GradAlign for training-free model performance inference.

    \item  GradAlign is efficient to compute and effective for  training-free inference of model performance at initialization. Through extensive experimentation on widely recognized NAS benchmarks, our results demonstrate that GradAlign generally surpasses the performance of existing training-free NAS methods.

     \item We finally investigate whether the number of linear regions \cite{montufar2014number,hanin2019complexity,serra2018bounding,zhang2020empirical} can be served as a strong indicator for model performance inference, which is the most commonly used metric in training-free NAS \cite{mellor2021neural,chen2021neural}. Our findings indicate that solely counting the number of linear regions may not be a reliable means of evaluating the network at initialization.
\end{itemize}
\vspace{-0.3cm}
\section{Related Works}

\begin{table*}[t]
\small
    \centering
    \begin{threeparttable}
    \begin{tabular}{|c|c|c|c|}
    \toprule
        Methodology & Inputs  &  Labels  & Gradients\\
        \midrule
       Number of Linear Regions  \cite{mellor2021neural,chen2021neural} & \cmark & \xmark &  \xmark\\
       Density of Sample-wise Losses \cite{zhang2021gradsign} & \cmark & \cmark & \cmark \\
       Stability of NTK Matrix  \cite{chen2021neural} & \cmark & \xmark &\cmark  \\
        Gaussian Complexity of Linear Function \cite{lin2021zen} & \xmark & \xmark & \cmark \\
         Saliency-based Approaches\tnote{*}  \cite{abdelfattah2021zero} & \cmark & \cmark & \cmark \\
         Correlation of Activations \cite{mellor2021neural} & \cmark & \xmark & \cmark \\
        \midrule
        Local Model Convergence (Ours) & \cmark & \cmark & \cmark \\
        \bottomrule
    \end{tabular}
    \begin{tablenotes}
        \footnotesize
        \item[*] synflow \cite{tanaka2020pruning} needs no data to compute score.
      \end{tablenotes}
    \end{threeparttable}
    \caption{A summary of existing  training-free NAS methods and their dependency on the dataset.}
    \label{tab:summary}
\end{table*}

\noindent \textbf{Neural Architecture Search.} The objective of Neural Architecture Search (NAS) is to effectively discover efficient neural network architectures \cite{zoph2016neural,liu2018darts,tan2019mnasnet}. However, the early NAS approaches \cite{zoph2016neural,2018Learning,2017Progressive,tan2019mnasnet} that were based on reinforcement learning still require high computational costs. To reduce the computational burden, recent work has focused on accelerating the search. This has resulted in a large number of gradient descent-based algorithms \cite{liu2018darts,2018ProxylessNAS,2019Densely} and one-shot-based algorithms \cite{2018Efficient,guo2020single,2020Single,2019FairNAS}. Some research is also directed towards finding NAS algorithms that do not require training to further reduce the computation. NASWOT \cite{mellor2021neural} is the first proposed algorithm that evaluates candidate architectures by the number of linear regions, while TE-NAS \cite{chen2021neural} builds on this by incorporating the number of linear regions and NTK to score the network. GradSign \cite{zhang2021gradsign} is another fast, no-training algorithm that only requires the gradients of a network evaluated at a random initialization state. Zen-NAS \cite{lin2021zen} is an advanced zero-shot approach to scoring network architectures. It is computed in a data-agnostic way and requires only knowledge of the input shapes of the network architecture for model ranking. Zero-Cost \cite{abdelfattah2021zero}, on the other hand, provides a set of zero-cost proxies, including Grad Norm, snip \cite{lee2018snip}, grasp \cite{wang2020picking}, synflow \cite{tanaka2020pruning}, fisher \cite{theis2018faster} and jacob\_cov \cite{mellor2021neural}. These zero-cost proxies predict the accuracy of network architectures using only a minibatch of training data.

\noindent \textbf{Deep Learning Theory.} There is a long history of investigating the theoretical properties of neural networks \cite{cybenko1989approximation,bartlett1996valid,anthony1999neural}. Previous works on neural network have relied on various complexity measures such as norms \cite{neyshabur2015norm,neyshabur2017exploring}, margins \cite{neyshabur2017exploring,bartlett2017spectrally,anthony1999neural}, VC-dimension \cite{anthony1999neural,harvey2017nearly}, and fat-shattering bounds \cite{anthony1999neural} to bound the generalization of neural networks. In recent years, a powerful tool called the ``neural tangent kernel'' (NTK) \cite{jacot2018neural} has been proposed to understand deep neural networks. In the NTK regime, the network parameters stay close to the initialization, and the neural network function evolves as a linear function of the network parameters. Recent works \cite{allen2019can,bai2019beyond,li2020learning,zou2021understanding} has studied the learning dynamics of neural networks beyond the NTK regime. In addition, many neural network pruning methods \cite{lee2018snip,wang2020picking,tanaka2020pruning,theis2018faster} leverage various saliency metrics for assessing the stability of the network. One of the inspirations for pruning is to evaluate the importance of parameters, which can be extended for evaluating the entire network \cite{abdelfattah2021zero,mei2019atomnas}.

\iffalse
\noindent \textbf{Deep Learning Theory.} There is a long history of investigating the theoretical properties of neural networks \cite{cybenko1989approximation,bartlett1996valid,anthony1999neural}. Previous works on neural network have relied on various complexity measures such as norms \cite{neyshabur2015norm,neyshabur2017exploring}, margins \cite{neyshabur2017exploring,bartlett2017spectrally,anthony1999neural}, VC-dimension \cite{anthony1999neural,harvey2017nearly}, and fat-shattering bounds \cite{anthony1999neural} to bound the generalization of neural networks. In recent years, a powerful tool called the ``neural tangent kernel'' (NTK) \cite{jacot2018neural} has been proposed to understand deep neural networks. In the NTK regime, the network parameters stay close to the initialization, and the neural network function evolves as a linear function of the network parameters. NTK has been leveraged to prove the convergence of deep neural networks \cite{jacot2018neural,li2018learning,du2018gradient,allen2019convergence} and to provide generalization guarantees \cite{allen2019learning,arora2019fine,arora2019exact,cao2019generalization}. Recent works \cite{allen2019can,bai2019beyond,li2020learning,zou2021understanding} has studied the learning dynamics of neural networks beyond the NTK regime. Training-free NAS methods leverage ideas from current deep learning theories to score neural networks at initialization.
\fi
\vspace{-0.4cm}
\section{Preliminary}
\vspace{-0.2cm}
\subsection{Deep Neural Networks}
A feed-forward neural network $f(x): X \rightarrow Y$ is a mapping from the input space $X$ to the label space $Y$. Typically, the network has a hierarchical structure consisting of $L$ layers. We assume that the input dimension of layer $l$ is $d_{l-1}$ and the input to layer $l$ is denoted as $a_{l-1}$. The output of layer $l$ is computed as $a_l = \sigma(W_l a_{l-1} + b_l)$, where $W_l$ is a $d_{l-1} \times d_l$ matrix and $b_l$ is a $d_l$-dimensional vector. Here, $\sigma$ is an activation function used to introduce nonlinearity in the network. We consider the rectified linear unit (ReLU) activation function, defined as $\textnormal{ReLU}(x) = \max{(x, 0)}$. By composing all the layers together, the output $a_{L}$ is computed as $a_{L} = W_L\sigma(W_{L-1}(\cdots \sigma(W_1 x + b_1)) + b_L$. The label $y$ can be computed as $\argmax_i a_{L,i}$. Modern neural networks can have a complex structure, including skip connections \cite{he2016deep} and batch normalization \cite{ioffe2015batch}. As a result, designing the optimal architecture for a given dataset is a challenging task.
\vspace{-0.3cm}
\subsection{Training-free NAS}
We adopt the evaluation protocol outlined in \cite{mellor2021neural} to score networks at initialization and compare the performance of various training-free NAS methods. Assuming a total of $S$ network architectures and $T$ datasets, we sample $n$ images from each dataset $D_t$ to form a probe set $P_n^t$. Given a training-free NAS method $\mathcal{M}$, we can compute the preference of dataset $D_t$ for architecture $A_s$ as follows: 
\begin{equation}
    r_{s,t} = \mathcal{M}(M_s, P_n^t)
\end{equation}
Intuitively, $r_{s,t}$ serves as indicator for the performance of architecture $A_s$ on the dataset $D_t$. Assume the ground-truth accuracy of the architecture $A_s$ on the dataset $D_t$ is denoted as $w_{s, t}$. The performance of the training-free NAS method $\mathcal{M}$ is measured via Kendall’s $\tau$ \cite{kendall1938new} between $ r_{s,t}$ and $w_{s, t}$ for all the architecture,
\begin{equation}
    \textnormal{score}(\mathcal{M}) = \frac{1}{T}\sum_{t=1}^T \tau(\{ r_{s,t}: s \in S \}, \{w_{s, t}: s \in S \} )
\end{equation}
%\textcolor{red}{TODO} where $[\alpha_{s,t}]$ and $[w_{s, t}]$ are lists of transferability estimation scores and ground-truth fine-tuning accuracies obtained by each pre-trained model $s$, respectively. 
A larger $\textnormal{score}(\mathcal{M})$ indicates that the estimation of the training-free NAS method accurately correlates with the final accuracy. Table \ref{tab:summary} shows a summary of existing training-free model performance inference methods based on the methodologies.

\vspace{-0.3cm}
\section{Theoretical Foundations}

\label{sec:theory}
\noindent \textbf{Insights.}  When applying gradient descent to optimize deep neural networks, a common approach involves minimizing the average loss computed across individual samples. For instance, given a dataset $\{(x_i, y_i)\}_{i=1}^N$ and a loss function $\ell$, the optimization goal is to minimize $L = \frac{1}{N}\sum_{i=1}^N\ell(f_\theta(x_i), y_i)$. \cite{zhang2021gradsign} observed that this loss function can be naturally decomposed into per-sample objectives. This insight serves as the motivation for GradSign which leverages the relative distance among sample-wise local minima for training-free performance inference. However, since only the initial parameter $\theta_0$ is accessible at initialization, GradSign needs to make the assumption that the per-sample local minima exist within a proximity of the initialization parameter. Nevertheless, the validity of this assumption has been brought into question by findings in \cite{allen2019can}, indicating that the optimization process for deep neural networks may not converge towards solutions of minimal complexity. To avoid depending on assumptions regarding the global convergence behavior of deep neural networks, our approach shifts away from analyzing per-sample local minima. Instead, we leverage per-sample gradients for analyzing local convergence behavior to enable training-free model performance inference.

\begin{figure}[!t]
   \centering
\begin{tabular}{cc}
\includegraphics[width=0.33\textwidth]{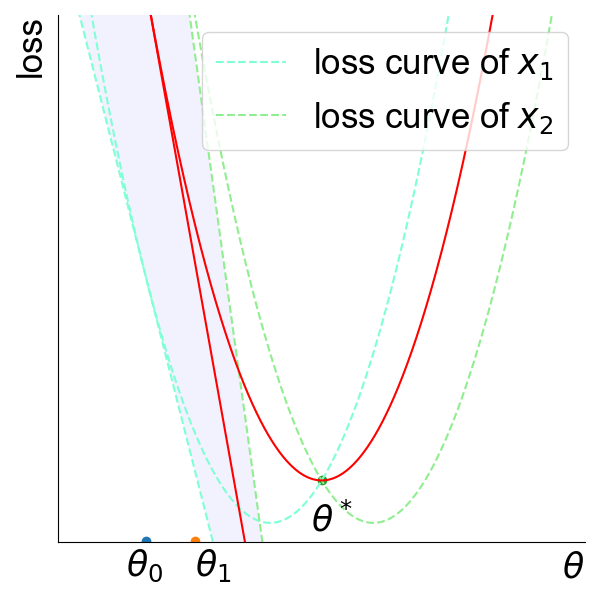} &
\includegraphics[width=0.33\textwidth]{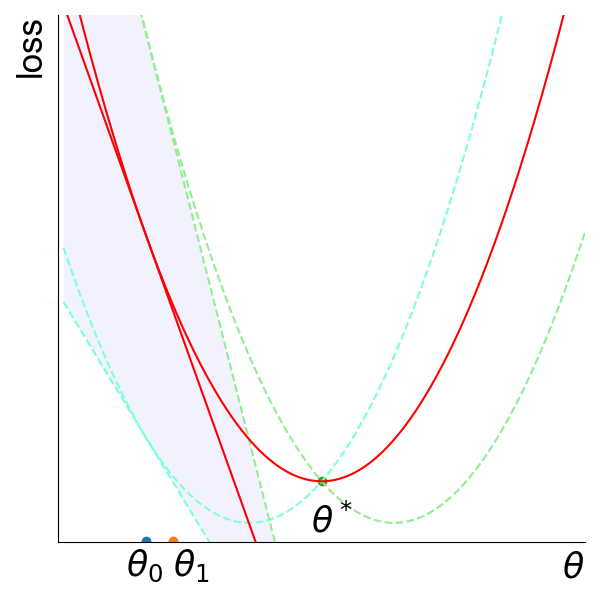} \\
a) & b)
\end{tabular}
    \caption{The network on the left is preferable, as the model can achieve faster convergence compared to the network on the right. The dotted lines are per-sample losses and gradients. The red lines are average losses and gradients. $\theta_0$ and $\theta_1$ are the initial parameters and the parameters after one-step gradient descent. For the left network, the updated parameters $\theta_1$ are closer to the optimal parameters $\theta^*$ in comparison to the network on the right. }
    \label{fig:depth}
\end{figure}

Consider two neural network architectures denoted as $f_\theta(x)$ and $f'_\theta(x)$, along with sample points $x_1$ and $x_2$ (Figure \ref{fig:depth}). To initiate our analysis, we evaluate the per-sample gradients at the initialization state for the network $f_\theta(x)$. Specifically, let $g_1 = \nabla_{\theta} \ell( f_{\theta}(x_1), y_1)|{\theta_0}$ and $g_2 =\nabla_{\theta} \ell( f_{\theta}(x_2), y_2)|{\theta_0}$. Similarly, for the network $f'_\theta(x)$, we denote the per-sample gradients as $g'_1$ and $g'_2$ for samples $x_1$ and $x_2$ respectively. We further introduce the angles $\beta$ and $\beta'$ as follows: $\beta = \arccos \frac{g_1 \cdot g_2 }{ \lVert g_1 \rVert \lVert g_2 \rVert}$ and $\beta' = \arccos \frac{g'_1 \cdot g'_2 }{ \lVert g'_1 \rVert \lVert g'_2 \rVert}$. Consider the scenario in which $\beta$ is significantly greater than  $\beta'$. For $\beta$, this implies a significant misalignment or conflict between the per-sample gradients, which can impede the convergence speed of the model. Consequently, the network $f'_\theta(x)$ is favored for its faster convergence. Interestingly, similar strategies are employed in continual learning, specifically for evaluating the interference among distinct tasks \cite{lopez2017gradient,chaudhry2018efficient,guo2020improved}. In what follows, we show a theoretical analysis of the impact of conflicting sample-wise gradients on the local model convergence.

\noindent \textbf{Theoretical analysis of conflicting sample-wise gradients.} We analyze the convergence behavior of conflicting sample-wise gradients in gradient descent. Following the above discussion, we consider two samples case $L = \ell_1 + \ell_2$, where $ \ell_1 = \ell(f_\theta(x_1), y_1)$ and  $\ell_2 = \ell(f_\theta(x_2), y_2)$, we have the following theorem,
\vspace{-0.1cm}
\begin{theorem}
   Assuming that $L$ is a differentiable function with a gradient that satisfies an $M$-Lipschitz condition ($M > 0$), the learning rate $\lambda$ satisfying $\lambda \le \frac{1}{M}$, and the per-sample gradient norm is bounded by $\sqrt{G}$. $\theta$ is the current model parameters and the $\theta^+$ is the model parameters after one-step gradient descent. The reduction in the loss function can be bounded as:
    \begin{align}
 L(\theta) -  L(\theta^+) \ge \frac{1}{2}\lambda (2G + \cos \beta G^2) 
\end{align}
\end{theorem}
\noindent Proof.  Let $\mathbf{g} = \mathbf{g}_1 + \mathbf{g}_2$, $\cos \beta = \frac{\mathbf{g}_1 \cdot \mathbf{g}_2}{ \rVert \mathbf{g}_1 \lVert \rVert \mathbf{g}_2 \lVert }$, then we have
\begin{align}
    L(\theta^+) & \le L(\theta + \nabla L(\theta)^T (\theta^+ - \theta) + \frac{1}{2}\nabla^2 L(\theta)\rVert \theta^+ - \theta \lVert^2 \\
    & \le L(\theta) + \nabla L(\theta)^T (\theta^+ - \theta)+ \frac{1}{2} M \rVert \theta^+ - \theta \lVert^2
\end{align}
Since $\theta^+ = \theta - \lambda \cdot  \mathbf{g}$, we have,
\begin{align}
    L(\theta^+) & \le L(\theta) - \lambda\mathbf{g}^T \cdot ( \mathbf{g}_1 + \mathbf{g}_2) + \frac{1}{2}M \lambda^2  \rVert \mathbf{g} \lVert^2 \\
    & \le L(\theta) - \lambda \rVert  \mathbf{g}_1 + \mathbf{g}_2 \lVert^2 + \frac{1}{2}M \lambda^2  \rVert \mathbf{g}_1 + \mathbf{g}_2\lVert^2 \\
    &  = L(\theta) - (\lambda - \frac{1}{2}M \lambda^2) \rVert  \mathbf{g}_1 + \mathbf{g}_2 \lVert^2 \\
   &  = L(\theta) - (\lambda - \frac{1}{2}M \lambda^2) ( \rVert \mathbf{g}_1 \lVert^2 + 2 \mathbf{g}_1 \cdot \mathbf{g}_2 +  \rVert \mathbf{g}_2 \lVert^2) \\
      &  = L(\theta) - (\lambda - \frac{1}{2}M \lambda^2) ( \rVert \mathbf{g}_1 \lVert^2 + \cos \beta \rVert \mathbf{g}_1 \lVert^2 \cdot \rVert \mathbf{g}_2 \lVert^2  +  \rVert \mathbf{g}_2 \lVert^2) 
\end{align}
If $\lambda \le \frac{1}{M}$, then $ - (\lambda - \frac{1}{2}M \lambda^2) \le -\frac{1}{2}\lambda$ and we have $L(\theta^+)  \le L(\theta) - \frac{1}{2}\lambda (2G + \cos \beta G^2)$. The theorem above demonstrates that a small value of $\beta$ results in a more significant reduction of the loss, consequently implying a faster convergence. Consequently, networks capable of generating low conflicting per-sample gradients are preferred.

\begin{algorithm}[!t]
\small
\caption{The proposed GradAlign for model performance inference at initialization.}
%the proposed algorithm for lifelong learning. $T$ is the number of tasks. $n_t$ is the number of mini-batches of task $t$. $M$ is the episodic memory. $\xi_k^t$ is the $k$-th mini-batch of task $t$ and $y_k^t$ is the corresponding label. $\zeta_k^t$ is a random mini-batch from the episodic memory. $w_k^t$ stands for the parameter after $k$-th mini-batch during the training of $t$-th task. $\ell_t(w_k^t;\xi_k^t)$ is the training loss calculated on $\xi_k^t$. $\ell_{\text{ref}}(w_k^t;\zeta_k^t)$ is the reference loss calculated on $\zeta_k^t$. $\alpha_1$ and $\alpha_2$ are defined in Equation \ref{eq:formulation2}.  }
\label{alg:Align}
\begin{algorithmic}[1]
\State Given a dataset $\{(x_i, y_i)\}_{i=1}^N$ and a model $f_\theta(x)$. $C$ is the total number of classes.
\State s = 0
\For{$c \gets 1$ to $C$}
\State  Gather all the samples with label $c$ as $\{(x_i, c)\}_{i=1}^{N_c}$. $N_c$ is the number of samples with label $c$.

\State Compute the per-sample gradients of all the samples at initialization and denoted as $\{g_i\}_{i=1}^{N_c}$.

\State \textbf{GradAlign-\rom{1}}: Compute the alignment score $s'$ based on Equation \ref{eq: 1}.
\State \textbf{GradAlign-\rom{2}}: Compute the alignment score $s'$ based on Equation \ref{eq: 2}.
\State $s$ = $s$ + $s'$.
\EndFor
\State $s = s / C$.
\State \Return $s$.
\end{algorithmic}
\end{algorithm}

\vspace{-0.3cm}
\section{GradAlign}
Based on our theoretical analysis, we introduce a straightforward yet effective approach, called GradAlign, for inferring model performance during initialization. Specifically, given a batch of image-label pairs $\{(x_i, y_i)\}_{i=1}^N$, our initial step involves segregating the samples based on their class labels, as it is important for ensuring minimal conflict within per-sample gradients of the same class. While it is possible measure the distinctness of the gradients among samples from different classes, our empirical findings indicate its comparatively limited efficacy. One plausible explanation is that the relationship between class-wise gradient similarity relies on the inherent interrelations among the classes themselves, which is ambiguous. Without losing generality, we assume the samples are from the same classes, and the GradAlign is determined by averaging the scores calculated across all the classes. The detailed algorithm of GradAlign is shown in Algorithm \ref{alg:Align}.

In GradAlign, we first compute the per-sample gradients of all the samples at initialization and denoted as $\{g_i\}_{i=1}^N$. Similar to \cite{balles2018dissecting} and \cite{bernstein2018signsgd}, we apply the sign function on the gradients to obtain the component-wise gradient direction as $\{\textnormal{sign} (g_i)\}_{i=1}^N$. We consider two strategies to capture the overall gradient alignments across the samples.
\\

\noindent \textbf{GradAlign-\rom{1}}. In this strategy, we first compute the mean gradient direction as $\Tilde{g} = \textnormal{sign} (\frac{1}{N} \sum_{i=1}^N g_i)$. Then we compute the alignment of per-sample gradients to the average gradient as,
\begin{equation}
   s=\frac{1}{N} \sum_{i=1}^{N}\textnormal{sign}(g_i)\cdot\Tilde{g}
    \label{eq: 1}
\end{equation}

A larger $s$ indicates that the per-sample gradients are more concentrated which can lead to larger loss reduction. Essentially, GradAlign-\rom{1} calculates the average pairwise alignment of gradient directions across all samples.
\\

\noindent \textbf{GradAlign-\rom{2}}. In the second strategy, we aim to measure the diversity of the per-sample gradients. In particular, we first compute the Gram matrix of the per-sample gradients,
\begin{equation}
{\centering
G = 
\begin{bmatrix}
    \textnormal{sign}(g_1) \cdot \textnormal{sign}(g_1) &   \dots  &  \textnormal{sign}(g_1) \cdot \textnormal{sign}(g_n) \\
   \vdots & \ddots & \vdots  \\
     \textnormal{sign}(g_n) \cdot \textnormal{sign}(g_1) &   \dots  &  \textnormal{sign}(g_n) \cdot \textnormal{sign}(g_n)
\end{bmatrix}}
\end{equation}
In order to capture the diversity of the per-sample gradients, we compute the log-determinant of the gram matrix as follows,
\begin{equation}
   s = \logdet G
       \label{eq: 2}
\end{equation}
The determinant of the Gram matrix is related to the volume spanned by the gradient vectors in the space. A large determinant indicates that the vectors span a relatively large volume, which implies that they are not concentrated in a lower-dimensional subspace. Hence, a lower value indicates that the per-sample gradient vectors are more tightly concentrated, suggesting that the model is likely to achieve faster convergence. 

Compared with GradSign \cite{zhang2021gradsign}, GradAlign does not require strong assumptions on the global convergence behavior of DNNs. Instead, GradAlign only relies on the per-sample gradients which local optimization behavior can be theoretically analyzed. Since the network can not be updated, we can only analyze the per-sample gradient at initialization, still empirically we found that they provide strong indication on the final performance of the model.

\vspace{-0.3cm}
\section{Experiments}

\subsection{Baselines}
We compare GradAlign with several currently popular methods for network architectures searching without training, including NASWOT \cite{mellor2021neural}, TE-NAS \cite{chen2021neural}, GradSign \cite{zhang2021gradsign}  and Zen-NAS \cite{lin2021zen}, on different benchmarks. In addition, we also include some Zero-Cost \cite{abdelfattah2021zero} methods, such as Grad Norm \cite{abdelfattah2021zero}, snip \cite{lee2018snip}, grasp \cite{wang2020picking}, synflow \cite{tanaka2020pruning}, fisher \cite{theis2018faster} and jacob\_cov \cite{mellor2021neural}. To ensure fairness, we used exactly the same experimental setup for all methods.

\subsection{Setups}

We follow the experimental setups as in \cite{mellor2021neural} for all the experiments. We utilize the default random initialization for every candidate architecture and subsequently feed a mini-batch of data into the network for testing. The batch size is fixed at 128, and no further data augmentation is applied to the input data. 
%\textbf{The code is attached in the supplementary material for reproducibility of all the experiments of the paper.}

\subsection{NAS Benchmarks}

To evaluate different approaches, we employ three widely recognized benchmarks: NAS201 \cite{dong2020bench}, NAS101 \cite{ying2019bench}, and NDS \cite{2019On}.

\noindent \textbf{NAS-BENCH-101} is the first large-scale public network architecture search dataset, containing 423k unique convolutional architectures in a compact search space and their results after multiple evaluations on the CIFAR-10 dataset. Thanks to the established benchmark, NAS algorithms can be compared quickly and rigorously without significant computational resources.

\noindent \textbf{NAS-BENCH-201} is designed as an extension of NAS-BENCH-101 to support multiple datasets and more diagnostic information. Its search space is composed of 15,625 neural cell candidates consisting of 4 nodes and 5 association operations. All candidate architectures are trained and evaluated on CIFAR-10, CIFAR-100 and ImageNet16-120 datasets in the same setting. Accuracy is also provided as additional diagnostic information.

\noindent \textbf{NAS DESIGN SPACE (NDS)} is a unique network architecture search framework. NDS is designed with the idea of comparing different search spaces rather than search algorithms. It has five different search spaces: DARTS \cite{liu2018darts}, NAS-NET \cite{2018Learning}, ENAS \cite{2018Efficient}, PNAS \cite{2017Progressive}, Amoeba \cite{2018Regularized}. All candidate architectures have been fully trained on CIFAR-10 dataset.It is worth noting that the network structures contained in each of the five search spaces described above have varying widths and depths. As a comparison, NDS additionally provides a fixed depth DARTS search space.

\subsection{Results}

\subsubsection{NAS-BENCH-101} 
The results in Table \ref{tab: 101_acc} shows that GradAlign-\rom{1} and GradAlign-\rom{2} outperform all other methods except Zen-NAS. In particular, compared with GradSign which also leverages gradients information for training-free model performance inference, GradAlign-\rom{1} and GradAlign-\rom{2} correlate better with model performance. In addition, while GradAlign-\rom{2} does not achieve the highest Kendall’s $\tau$ score, it picks out the relatively best performed networks (Table \ref{tab: 101_acc}). This is critical for real-world applications of training-free NAS as we often just select the top-ranked architecture for model training.

\begin{table*}[!h]
\caption{Results of Kendall’s $\tau$ on NAS Bench 101. While GradAlign-\rom{1} and GradAlign-\rom{2} achieve lower Kendall’s $\tau$ compared with Zen-NAS, they are better than other baselines.   }
\centering
\resizebox{\textwidth}{!}{
\begin{tabular}{l|cccccccccccc}
\toprule
Methods &  GradAlign-\rom{1}& GradAlign-\rom{2} &GradSign & NASWOT & TE-NAS &  fisher & grad\_norm & grasp & jacob\_cov  & snip & Synflow &Zen-NAS\\
\midrule
Kendall’s  $\tau$&0.287 & 0.284 &0.278 & 0.244 &  0.271 & -0.247  & -0.207 & 0.175  & -0.029 & -0.186 &0.254& \textbf{0.395} \\
\bottomrule
\end{tabular}}
\label{tab: 101_tau}
\end{table*}

\begin{table*}[!h]
\caption{The accuracy of top-ranked network on NAS Bench 101. }
\centering
\resizebox{\textwidth}{!}{
\begin{tabular}{l|cccccccccccc}
\toprule
Methods &  GradAlign-\rom{1}& GradAlign-\rom{2} &GradSign & NASWOT & TE-NAS &  fisher & grad\_norm & grasp & jacob\_cov  & snip & Synflow &Zen-NAS\\
\midrule
Accuracy (\%) &93.199 & \textbf{93.630} &93.609 & 93.098 &  89.873 & 42.347  & 84.104 & 80.639  & 90.584 & 84.104 &89.052& 91.957\\
\bottomrule
\end{tabular}}
\label{tab: 101_acc}
\end{table*}
\begin{table}[!h]
\caption{Results of Kendall’s $\tau$ on NAS Bench 201. GradAlign-\rom{1} and GradAlign-\rom{2} achieve higher Kendall’s $\tau$ than other baselines.   }
\centering
\begin{tabular}{l|ccc}
\toprule
Methods & CIFAR10 & CIFAR100 & ImageNet16-120\\
\midrule
GradAlign-\rom{1} & \textbf{0.617} & \textbf{0.633} & \textbf{0.623}\\
GradAlign-\rom{2} & 0.599 & 0.622 & 0.607\\
\midrule
GradSign  & 0.561 & 0.595 & 0.596  \\
NASWOT  & 0.587 & 0.621 & 0.594  \\
TE-NAS  &0.387 & 0.377 & 0.345  \\
fisher  &0.377 & 0.404 & 0.367  \\
grad\_norm & 0.437 & 0.464 & 0.427  \\
grasp &0.319 & 0.379 & 0.411  \\
jacob\_cov&0.553 & 0.549 & 0.554  \\
%plain &-0.175 & -0.287 & -0.145  \\
snip &0.435 & 0.466 & 0.432  \\
Synflow & 0.540 & 0.568 & 0.561  \\
Zen-NAS&0.180 & 0.188 & 0.219  \\
\bottomrule
\end{tabular}
\label{tab: 201_tau}
\end{table}

\begin{table}[!h]
\caption{The accuracy (\%) of top-ranked network on NAS Bench 201. }
\centering
\begin{tabular}{l|ccc}
\toprule
Methods & CIFAR10 & CIFAR100 & ImageNet16-120\\
\midrule
GradAlign-\rom{1} & \textbf{90.223} & \textbf{71.106} & \textbf{41.444}\\
GradAlign-\rom{2} & \textbf{90.223} & \textbf{71.106} & \textbf{41.444}\\
\midrule
GradSign  & \textbf{90.223} & \textbf{71.106} & \textbf{41.444} \\
NASWOT  & 85.937 & 68.940 & 35.478  \\
TE-NAS  &10.000 & 1.000 & 0.833  \\
fisher  &85.12 & 62.419 & 29.266  \\
grad\_norm & 85.120 & 62.420 & 5.456  \\
grasp &84.890 & 47.620 & 31.333  \\
jacob\_cov&87.030 & 69.800 & 36.933  \\
%plain &-0.175 & -0.287 & -0.145  \\
snip &85.120 & 62.420 & 5.456  \\
Synflow & \textbf{90.223} & \textbf{71.106} & \textbf{41.444}  \\
Zen-NAS&86.960 & 68.260 & 40.600  \\
\bottomrule
\end{tabular}
\label{tab: 201_acc}
\end{table}

\begin{table*}[!h]
\caption{Results of Kendall’s $\tau$ on NDS. GradAlign-\rom{1} achieves higher Kendall’s $\tau$ than other baselines. }
\centering
\begin{tabular}{l|ccccccc}
\toprule
Methods & Amoeba& DARTS & DARTS-fix-w-d& ENAS & NASNet& PNAS\\
\midrule
GradAlign-\rom{1} & \textbf{0.274} & \textbf{0.545} & \textbf{0.189} & \textbf{0.444} & \textbf{0.306} & \textbf{0.415} \\
GradAlign-\rom{2} & 0.257  & 0.540  & 0.148 & 0.425 &  0.298&  0.408\\
\midrule
GradSign  & 0.260 & 0.541 & 0.153 & 0.427 & 0.293 & 0.408  \\
NASWOT  & 0.187 & 0.462 & 0.102 & 0.370 & 0.278 & 0.360 \\
TE-NAS  & 0.039 & 0.164 & 0.021 & 0.205 & -0.533 & 0.151  \\
fisher & -0.122 & 0.197 & -0.213 & 0.032 & -0.099 & 0.040  \\
grad\_norm  & -0.113 & 0.228 & -0.186 & 0.056 & -0.078 & 0.097  \\
grasp  & 0.072 & -0.052 & 0.086 & 0.053 & 0.171 & 0.021 \\
jacob\_cov  & 0.236 & 0.192 & 0.169 & 0.122 & 0.122 & 0.158  \\
snip  & -0.079 & 0.272 & -0.157 & 0.093 & -0.048 & 0.152  \\
Synflow  & -0.053 & 0.299 & -0.092 & 0.133 & 0.019 & 0.181  \\
Zen-NAS  & -0.020 & 0.323 & 0.011 & 0.183 & 0.071 & 0.187 \\
\bottomrule
\end{tabular}
\label{tab: nds_tau}
\end{table*}

\subsubsection{NAS-BENCH-201}
As shown in Table \ref{tab: 201_tau}, GradAlign-\rom{1} and GradAlign-\rom{2} are the best performing algorithms in all three NAS-BENCH-201 datasets. In particular, GradAlign-\rom{1} outperforms NASWOT, the best performing algorithm among the other algorithms used as comparisons, by an average of 4\% over the three datasets. Meanwhile, both GradAlign-\rom{1} and GradAlign-\rom{2} select the relatively best networks as shown in Table \ref{tab: 201_acc}.
\subsubsection{NAS DESIGN SPACE}
We conducted experiments in all six search spaces, and all results are summarized in Table \ref{tab: nds_tau} and Table \ref{tab: nds_acc}. Table \ref{tab: nds_tau} shows that GradAlign-\rom{1} achieves the highest the highest Kendall’s $\tau$ score in all the search spaces while GradAlign-\rom{2} also performs competitively across all the cases. However, as search spaces are highly diverse, there is no single method which always picks out the best performed network. Nevertheless, both GradAlign-\rom{1} and GradAlign-\rom{2} identify a well-performing network that is in closer proximity to the best-performing network.

Table \ref{tab: rankings} presents a summary of method performance across all benchmarks, as measured by Kendall’s $\tau$ score. Overall, it is evident that GradAlign-\rom{1} and GradAlign-\rom{2} secure the first and second ranks, respectively. Furthermore, depending on the specific benchmark, GradAlign-\rom{2} excels in identifying the best performing network.
\vspace{-0.3cm}
\section{Running Time Analysis}
\vspace{-0.2cm}
On CIFAR10, using NAS-BENCH-201, the average search time for GradAlign-\rom{1} is 2.93s (reduced to \textbf{0.60s} by leveraging functools \footnote{https://docs.python.org/3/library/functools.html} for computing per-sample gradients in parallel). Meanwhile, NASWOT takes 0.80s, GradSign 2.35s, Synflow 1.19s, and TE-NAS 1.39s. These measurements were obtained with a batch size of 32 on an RTX 2080ti GPU. Consequently, GradAlign-\rom{1} exhibits superior computational efficiency compared to other methods.

\begin{table*}[!t]
\caption{The accuracy (\%) of top-ranked network on NDS. }
\centering
\begin{tabular}{l|ccccccc}
\toprule
Methods & Amoeba& DARTS & DARTS-fix-w-d& ENAS & NASNet& PNAS\\
\midrule
GradAlign-\rom{1} & 93.640 & \textbf{94.360} & 92.170 & 93.670 & 92.630 & 93.380 \\
GradAlign-\rom{2} & 93.640  & 93.530  & 92.170 & \textbf{94.160} & 92.630& 93.380\\
\midrule
GradSign  &  93.640 & 93.530 & 92.170 & \textbf{94.160} & 92.630 & 93.380 \\
NASWOT  & 93.320 & 93.530 & \textbf{92.900} & \textbf{94.160} & 94.250 & \textbf{94.670} \\
TE-NAS  & 90.850 & 92.410 & 91.480 & 92.680 & 93.170 & 92.110  \\
fisher & 91.830 & 87.480 & 10.000 & 83.310 & 92.210 &92.860\\
grad\_norm  & 91.830 & 87.480 & 10.000 & 83.310 & 90.190 & 88.650  \\
grasp  & 10.000 & 88.770 & 83.520 & 36.480 & 88.830 & 91.310 \\
jacob\_cov  & \textbf{93.960} & 93.920 & 87.620 & 93.530 & \textbf{94.740} & 94.260   \\
snip  & 91.830 & 87.480 & 10.000 & 83.310 & 90.390 & 88.650 \\
Synflow  & 90.830 & 93.710 & 10.000 & 93.350 & 92.930 & 91.580  \\
Zen-NAS  & 91.630 & 92.670 & 92.170 & 93.020 & 78.410 & 93.000 \\
\bottomrule
\end{tabular}
\label{tab: nds_acc}
\end{table*}

\begin{table}[!t]
\caption{The average ranking of the methods on all the benchmarks. GradAlign-\rom{1} and GradAlign-\rom{2} rank the first and the second seperately.  }
\centering
\begin{tabular}{l|ccc|c}
\toprule
Methods & NAS Bench 101 & NAS Bench 201 & NDS &MEAN \\
\midrule
GradAlign-\rom{1} & 2.000 & 1.000 & 1.000 & 1.333\\
GradAlign-\rom{2} & 3.000 & 2.000 & 2.833&2.611\\
\midrule
GradSign  & 4.000 & 3.667 & 2.333 & 3.333 \\
NASWOT  & 7.000 & 3.333 & 4.167 & 4.833 \\
TE-NAS  &5.000 & 10.333 & 8.333  & 7.889\\
fisher  &12.000 & 9.667 & 11.000 & 10.889\\
grad\_norm & 11.000 & 7.667 & 9.833  & 9.500\\
grasp &8.000 & 10.000 & 8.500  & 8.833\\
jacob\_cov&9.000 & 5.667 & 6.000 & 6.889 \\
%plain &-0.175 & -0.287 & -0.145  \\
snip &10.000 & 7.333 & 8.667  &8.667\\
Synflow & 6.000 & 5.333 & 7.333  & 6.222\\
Zen-NAS&1.000 & 12.000 &6.333 & 6.444 \\
\bottomrule
\end{tabular}
\label{tab: rankings}
\end{table}
\vspace{-0.3cm}
\section{Visualization}
\begin{figure}[!t]
   \centering
\begin{tabular}{ccc}
\includegraphics[width=0.29\textwidth]{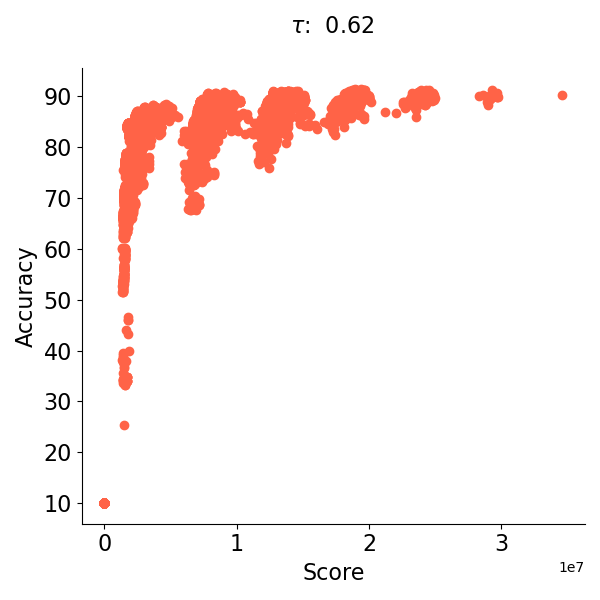} &
\includegraphics[width=0.29\textwidth]{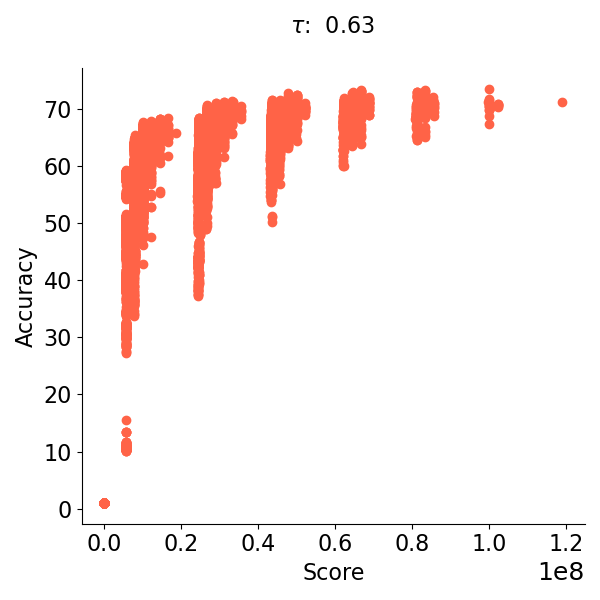} &
\includegraphics[width=0.29\textwidth]{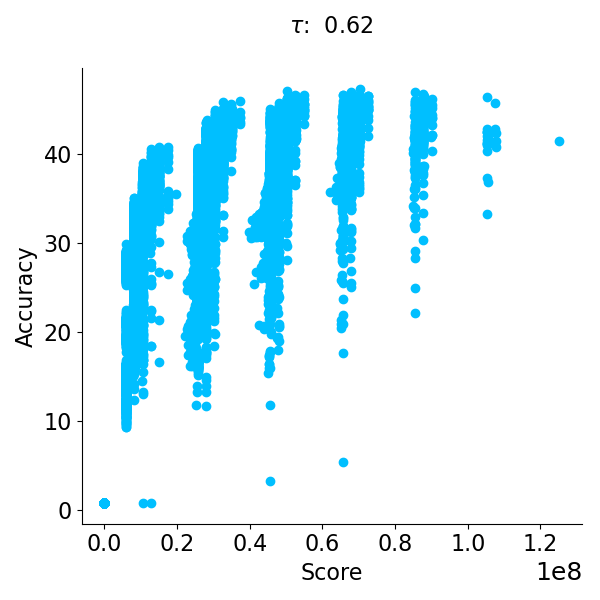} \\
a) & b) & c)
\end{tabular}
    \caption{Visualization of model testing accuracy versus GradAlign-\rom{1} metric score on CIFAR10,
CIFAR100, ImageNet16-120.}
    \label{fig:1}
\end{figure}

\begin{figure}[!t]
   \centering
\begin{tabular}{ccc}
\includegraphics[width=0.29\textwidth]{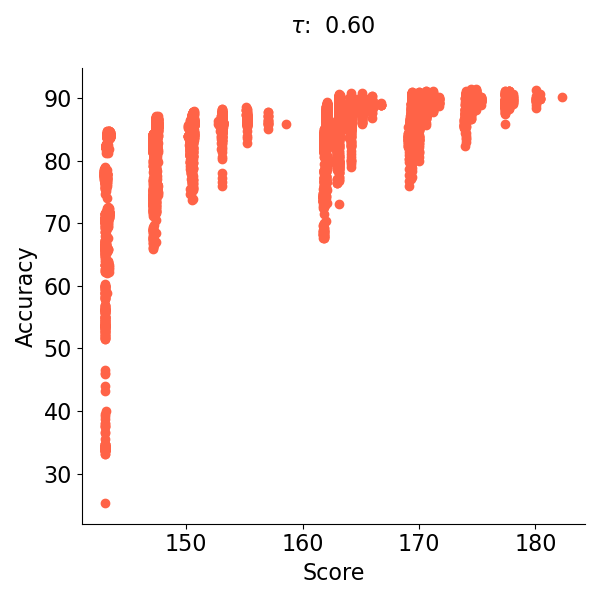} &
\includegraphics[width=0.29\textwidth]{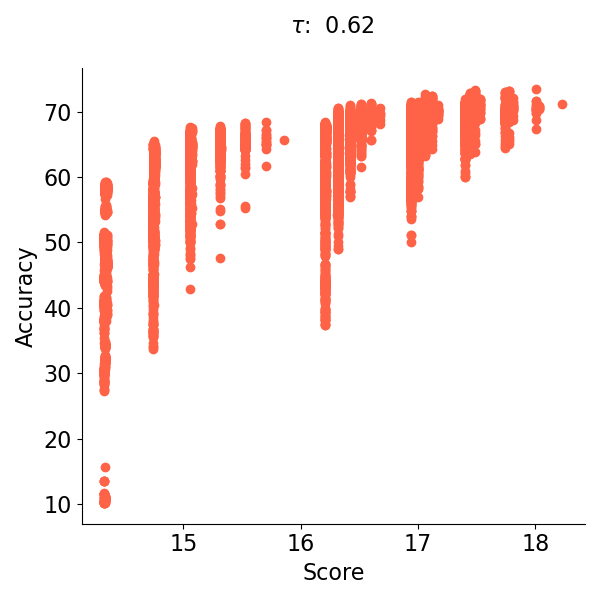} &
\includegraphics[width=0.29\textwidth]{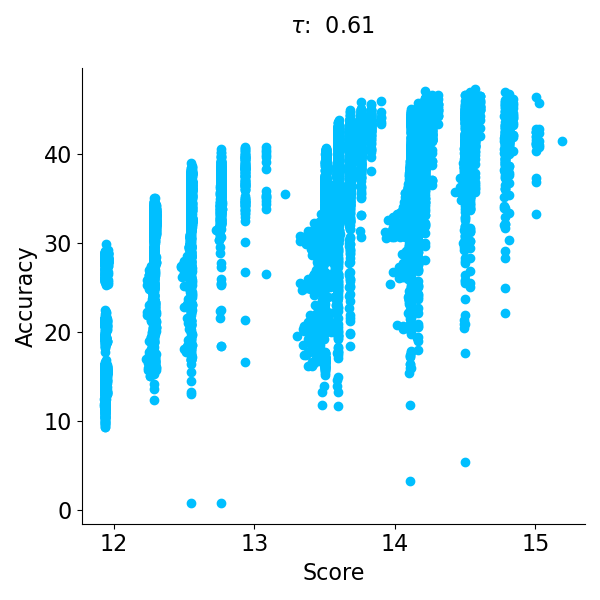} \\
a) & b) & c)
\end{tabular}
    \caption{Visualization of model testing accuracy versus GradAlign-\rom{2} metric score on CIFAR10,
CIFAR100, ImageNet16-120.}
    \label{fig:2}
\end{figure}

\vspace{-0.3cm}
The visualizations of model testing accuracy versus GradAlign-\rom{1} metric score and GradAlign-\rom{2} metric score on CIFAR10,
CIFAR100, ImageNet16-120 are shown in Figure \ref{fig:1} and Figure \ref{fig:2}, respectively. We also show the Kendall’s $\tau$ in each case. It can be observed that the GradAlign-\rom{1} metric score and the GradAlign-\rom{2} metric score have strong correlation with the model accuracy across all the datasets. This visualization further shows that GradAlign-\rom{1} metric score and GradAlign-\rom{2} can be used for indicating the model performance without training.

\section{The Number of Linear Regions is Sensitive to Network Initialization}
\label{sec: sensitive}

Finally, we present an analysis of a widely used metric for inferring training-free model performance. Specifically, we demonstrate that relying solely on the number of linear regions, as utilized in \cite{mellor2021neural} and \cite{chen2021neural}, may not provide a dependable indicator for assessing training-free model performance.

To begin with, we showcase the dependence of the number of linear regions on the initial values of the parameters. To demonstrate this, we construct a fully-connected neural network comprising three layers with ReLU activations and weight matrices $W_1 \in \mathbb{R}^{2 \times 2}$, $W_2 \in \mathbb{R}^{2 \times 2}$, and $W_3 \in \mathbb{R}^{2 \times 1}$, in addition to biases $B_1 \in \mathbb{R}^{2 \times 1}$, $B_2 \in \mathbb{R}^{2 \times 1}$, and $B_3 \in \mathbb{R}$. For a given input $x$, the output of the first layer is computed as $ A_1 = \max(W_1x+B_1, 0)$, the output of the second layer is computed as $ A_2 = \max(W_2A_1+B_2, 0)$, and the final output is computed as $O = \max(W_3A_2+B_3, 0)$. This example serves to demonstrate that slight perturbations in the parameter values can significantly alter the number of linear regions in the network.

\begin{figure}[!t]
   \centering
\setlength{\tabcolsep}{1.2pt}
\renewcommand{\arraystretch}{1.2}
\begin{tabular}{cccc}
   \includegraphics[align=c,height=0.12\textheight]{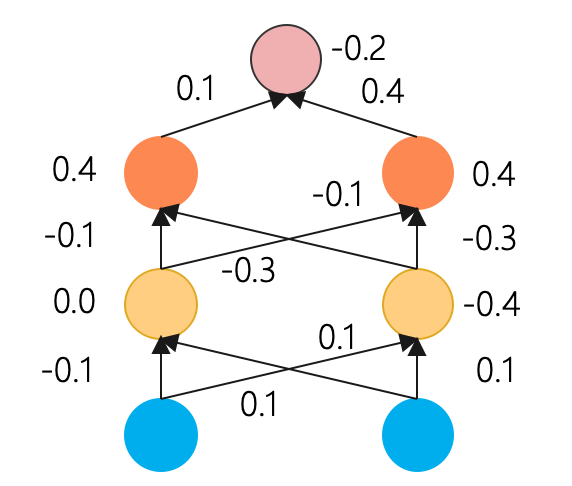} &
   \includegraphics[align=c,height=0.14\textheight]{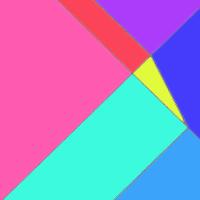} &
\includegraphics[align=c, height=0.12\textheight]{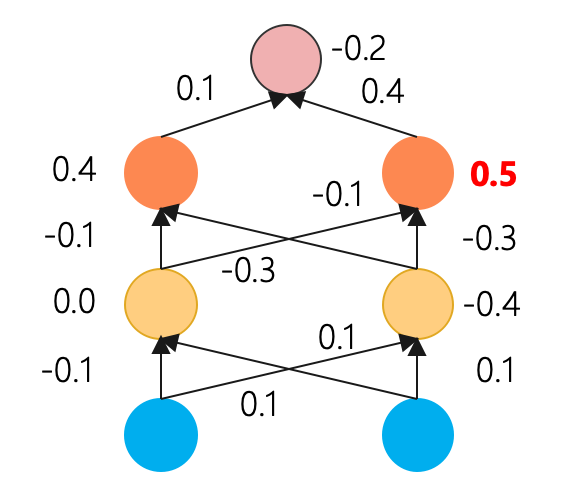} &
   \includegraphics[align=c, height=0.14\textheight]{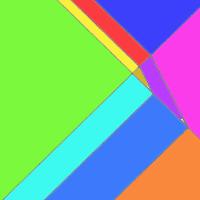} \\
   \multicolumn{2}{c}{a)} &  \multicolumn{2}{c}{b)} \\
\end{tabular}
    \caption{The number of linear regions is sensitive to the parameter values. By slightly perturbing the value of one parameter (marked in red), the number of linear regions greatly increases. }
    \label{fig:linear_regions}
\end{figure}

Figure \ref{fig:linear_regions} (a) depicts a random initialization of the network's parameters and the corresponding visualization of the hyperplanes partitioning the input space into 7 linear regions. The values adjacent to the nodes represent the initial bias values. In Figure \ref{fig:linear_regions} (b), we slightly perturb one of the parameters (marked in red), and the number of linear regions increases to 10. Through this simple example, we highlight the sensitivity of the number of linear regions to perturbations in network parameters. This finding suggests that relying solely on the number of linear regions for training-free NAS may not be entirely dependable. We note that similar findings have been reported in previous studies \cite{telgarsky2015representation,hanin2019complexity} examining sawtooth functions. Specifically, slight perturbations in the weights and biases of a ReLU network representing a sawtooth function can significantly decrease the number of linear regions.

In contrast, techniques that harness gradient information, such as GradSign \cite{zhang2021gradsign}, often attain superior performance when compared with approaches reliant on quantifying the number of linear regions. GradAlign follows a similar path by quantifying the divergence within per-sample gradients, which proves to be more dependable and indicative for inferring training-free model performance.

\vspace{-0.3cm}
\section{Conclusion}
\vspace{-0.2cm}
In this paper, we propose a simple yet effective method, call GradAlign, for training-free model performance inference. The proposed method is based on our theoretical analysis on the conflicts of per-sample gradients which indicates that larger conflicts can lead to slower model converge. GradAlign is based on a simple criterion that model with smaller per-sample gradients conflicts should be preferred. Extensive experiments show that GradAlign outperforms existing training-free NAS methods on several commonly used NAS benchmarks on average. Finally, we provide additional analysis showing that the number of linear regions may not be a reliable criterion for training-free NAS. Still, our findings indicate that, owing to the diversity among NAS benchmarks, no single approach attains optimal results in both Kendall’s $\tau$ score and the accuracy of top-ranked networks across all the cases. As an interesting and challenging future direction, we will investigate more stable training-free model inference methods which can consistently perform well across the cases.

\bibliographystyle{splncs04}
\bibliography{main}
\end{document}